%% file: main.tex
\documentclass[letterpaper, 10 pt, conference]{ieeeconf}  

\IEEEoverridecommandlockouts                              

\usepackage{graphics} 
\usepackage{tikz}
\usepackage{tkz-tab}
\usepackage{caption}
\usepackage{latexsym}
\usepackage{amssymb}
\usepackage{amsmath}
\usepackage{subcaption}
\usepackage{hyperref}

\title{\LARGE \bf
Can you text what is happening? 
Integrating pre-trained language encoders
into trajectory prediction models for autonomous driving}

\author{Ali Keysan$^{1,2}$, Andreas Look$^{1}$, Eitan Kosman$^{3}$, 
Gonca G\"ursun$^{1}$,
Jörg Wagner$^{1}$, Yu Yao$^{1}$, Barbara Rakitsch$^{1}$
\thanks{$^{1}$ The authors are with the Bosch Center for Artificial Intelligence, Renningen, Germany.}
\thanks{$^{2}$ The author is with the University of Tübingen, Tübingen, Germany.}
\thanks{$^{3}$ The author is with the Bosch Center for Artifical Intelligence, Haifa, Israel.}
}

\begin{document}

\maketitle
\thispagestyle{empty}
\pagestyle{empty}

\input{sections/01_abstract}
\input{sections/02_intro}
\input{sections/03a_background}
\input{sections/03b_methods}

\input{sections/03c_representation}
\input{sections/04_experiments}
\input{sections/05_summary}
\input{sections/06_outlook}
\input{sections/07_appendix}

\input{sections/08_acknowledgements}

\bibliography{sample}
\bibliographystyle{plain}

\end{document}

%% file: sections/01_abstract.tex
\begin{abstract}
In autonomous driving tasks, scene understanding is the first step towards predicting the future behavior of the 
surrounding traffic participants. Yet, how to represent a given scene and extract its features are still open research questions. 
In this study, we propose a novel text-based representation of traffic scenes and process it with a pre-trained language encoder.

First, we show that text-based representations, combined with classical rasterized image representations, lead to descriptive scene embeddings. 
Second, we benchmark our predictions on the nuScenes dataset and show significant improvements compared to baselines.
Third, we show in an ablation study that a joint encoder of text and rasterized images outperforms the individual encoders confirming that both representations have their complementary strengths. 
\end{abstract}

%% file: sections/02_intro.tex
\section{INTRODUCTION}
\label{sec:intro}



\textit{Autonomous driving} (AD) is a complex endeavor, which involves multiple tasks. 
One such task is behavior prediction, which deals with the problem of estimating future trajectories of the traffic participants surrounding the autonomous vehicle. 
Behavior prediction is a crucial component of an AD pipeline since it enables a number of downstream components such as motion planning to generate safe trajectories and avoid collisions.

On a different note, much progress has been made in recent years on \textit{foundation models} (FM). 
These are general purpose models which are pre-trained on large scale data, often in an unsupervised fashion, on a task that is agnostic to any specific application.
Within the domain of FMs, \textit{large language models} (LLM) are models 
 that are pre-trained on a large scale text corpus and whose pre-training task is to predict the next token or randomly masked tokens within a sequence.
LLMs have received significant attention, since they exhibit emergent capabilities. Despite their application agnostic training objective, they can solve a wide range of \textit{natural language processing} (NLP) tasks in a zero/few shot setting, e.g., via prompt engineering \cite{xie2022explanation}.
Due to their versatility, LLMs have also been applied to fields outside of NLP, such as time-series data \cite{xue2022prompt} or tabular data \cite{hegselmann2023tabllm}.


FMs have also found their way into AD in the form of neural simulators. These models use vision FMs to generate sensor data for a fictitious traffic scene, allowing for simulation of edge cases \cite{yang2023unisim}.
However, despite their popularity, language models have, to the best of our knowledge, not been applied as scene encoders for trajectory prediction tasks.
We hypothesize that this is due to the inherently complex nature of traffic scenes, which does not lend itself to a text-based representation.
And, unlike images, text is not a common input (or output) modality in AD tasks.
Furthermore, practical challenges, such as limited context length of language models, present additional obstacles.

In this work, we are the first to study the usage of text descriptions combined with pre-trained language encoders for the AD trajectory prediction task.
Our descriptions contain information about the agent state, its history and road lanes.
For the latter, a standard representation in form of polylines does often not fit into a single prompt, and we propose instead to encode their information in form of Bézier curves.
We evaluate our new approach on the nuScenes dataset \cite{caesar2020nuscenes} 
resulting in two key observations:
\begin{enumerate}
\item
We show that text descriptions combined with a pre-trained language encoder provide a viable alternative to rasterized images.
\item 
We show that image and text encoders have complementary
strengths and a joint encoder outperforms its individual counterparts.
\end{enumerate}
Our study serves as a proof-of-concept, showing that AD tasks can benefit from text-based representations.
We anticipate that our findings will trigger further research towards more interpretable and expressive prediction models.

The remaining paper is structured as follows: 
In Section~\ref{sec:background}, we introduce CoverNet \cite{phan2020covernet}, the baseline trajectory prediction model, which our architecture is based on. 
In Sections~\ref{sec:methods} and \ref{sec:representation}, we introduce our novel model architecture and the corresponding scene representation, respectively.
In Section~\ref{sec:experiments}, we empirically validate our approach.
Finally, in Sections~\ref{sec:summary} and \ref{sec:outlook}, we summarize our results and give an outlook on future work.

%% file: sections/03a_background.tex
\section{Background}
\label{sec:background}
Our approach takes a standard trajectory prediction model, CoverNet\cite{phan2020covernet}, and integrates pre-trained language encoders into the model architecture.
In the following, we provide further background information on the CoverNet architecture.

\begin{figure*}
    \centering
\includegraphics[width=0.85\linewidth]{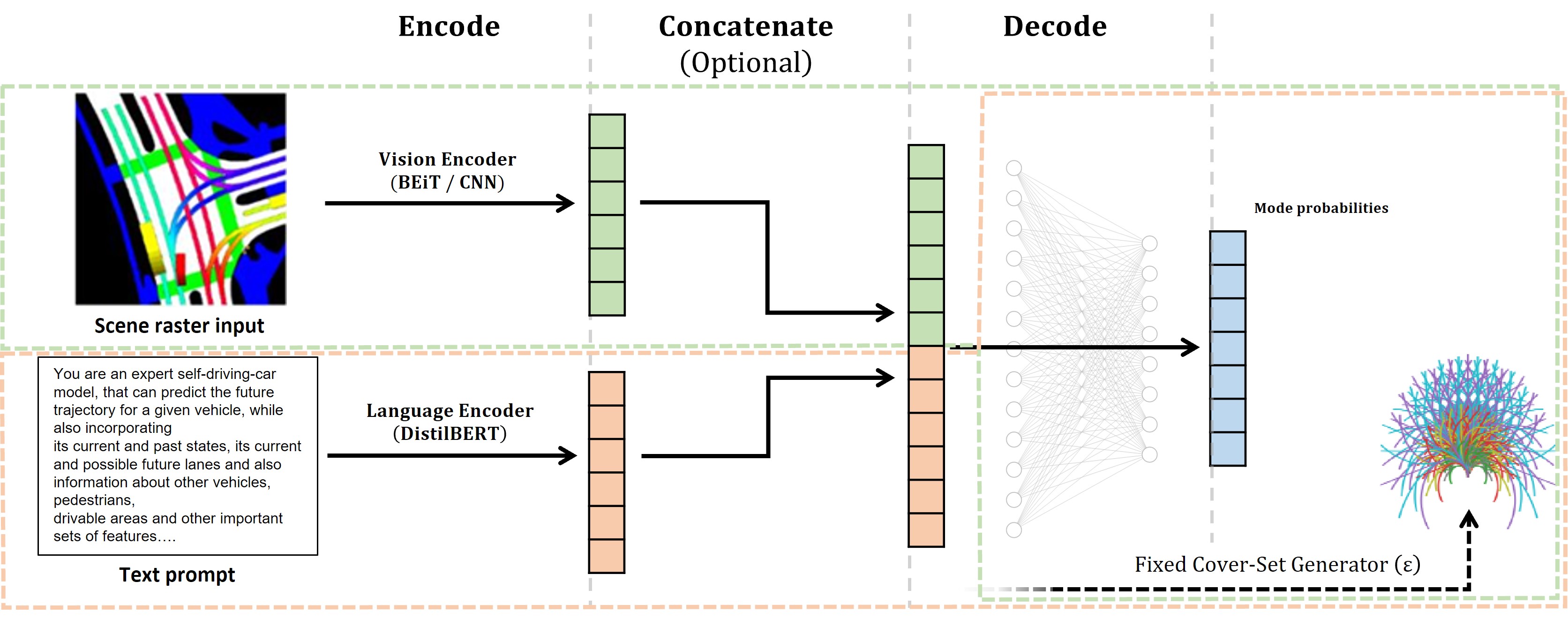}
 \caption{\textbf{Flow of our Model.} 
 We encode the image that represents the rasterized scene and the text prompt with pre-trained models dedicated for each modality. 
 If both input sources are used, we afterwards concatenate their embeddings.
 The result is fed into a decoder whose final layer picks the target trajectory from a pre  generated trajectory set.
 }
\label{fig:architecture}
\end{figure*}

\subsection{CoverNet}
\label{subsec:covernet}
CoverNet applies an encoder-decoder architecture that is fed with an image representation of the scene, processes it with a vision encoder backbone, and finally decodes the targets from the learned representation.

Similar to previous vision-based works, e.g. \cite{chai2019multipath, cui2019multimodal}, CoverNet encodes scenes into rasterized images
which are subsequently fed into a ResNet-50~\cite{he2016deep} architecture, a widely used variant of \textit{Convolutional Neural Networks} (CNNs). 
After extracting the feature map of an intermediate layer and applying global pooling, the resulting embedding is concatenated with an agent state vector and propagated through a series of dense layers in order to generate predictions.

The decoder treats the trajectory prediction task as a classification problem.
This is done by defining a set of trajectory candidates $\mathcal{K}$ from which the predicted trajectories are picked. 
Such a set has to cover a wide range of possible trajectories in order to contain a hypothesis that is close to the ground-truth.
At the same time, it is desired to keep the set as small as possible in order to avoid making the classification problem unnecessarily hard. 
An optimal cover-set can be found by first defining a tolerance parameter $\varepsilon>0$, and then searching for the minimal trajectory set $\mathcal{K}$ that covers all trajectories from a predefined trajectory set $\mathcal{K}'$:
$\forall k \in \mathcal{K}, l \in \mathcal{K}': \delta(k,l) < \varepsilon$, 
where $\delta$ is a metric that quantifies the distance between two trajectories.
While this problem is in general NP-hard, the authors show that simple, greedy solutions work well in practice \cite{phan2020covernet}.

%% file: sections/03b_methods.tex
\section{Model Architecture}
\label{sec:methods}

Our model follows a standard encoder-decoder architecture. We explore three options for encoders: image-based, text-based and a joint encoder, which we present in Sections~\ref{subsec:image_encoder}, \ref{subsec:text_encoder}, and \ref{subsec:fused_model}, respectively.
For the decoder, we follow the approach from CoverNet and treat the trajectory prediction task as a classification problem.
Figure~\ref{fig:architecture} shows an overview of our model architecture.

\subsection{Image Encoder}
\label{subsec:image_encoder}

The image encoder reduces the dimensionality of the high-dimensional image input by extracting relevant information into a lower-dimensional feature space.
CoverNet relies on a CNN-based encoder, which is the traditional model choice. 

Since the focus of our research is to assess the performance differences between the two input modalities image and text, we further expand our study to transformer-based image encoders.
Text-based encoders commonly employ transformers and this strategic choice allows us a fair comparison between the modalities that is not confounded by differences based on architectures.

\textbf{Vision Transformers} (ViTs) \cite{dosovitskiy2020image} adapt the original Transformer architecture \cite{vaswani2017transformer}, which was initially designed for language modeling tasks, to be applicable to images. 
This allows ViTs to effectively process and represent image data, challenging the conventional use of CNNs as the primary image encoder \cite{han2023vitsurvey}.
ViTs divide the input image into smaller patches, treating them as analogous to tokens in natural language processing. 

\textbf{Bidirectional Encoder Image Transformer} (BEiT) \cite{bao2021beit} is a state-of-the-art ViT, which we chose for our experiments.
It is pre-trained on ImageNet \cite{russakovsky2015imagenet}
and its pretraining objective involves reconstructing randomly masked image patches.

\subsection{Text Encoder}
\label{subsec:text_encoder}
The text encoder ingests a text-based representation of the scene and outputs a scene embedding which is subsequently decoded into a trajectory prediction.
In order to achieve this, a high-performance language model can be used. However, the large size of state-of-the-art language models are challenging for fine-tuning since they do not fit into memory. Hence, we select DistilBERT \cite{sanh2019distilbert} for our experiments, which is a slim variant of BERT \cite{devlin2018bert} that does not suffer from this issue.

\textbf{BERT}
is a transformer-based language model pre-trained via a masked language model objective, which requires the model to predict randomly masked out input tokens \cite{devlin2018bert}. 
In contrast to state-of-the-art pre-trained language models, which typically use an autoregressive architecture, BERT is bidirectional, i.e., every token can attend to both past and future tokens.  In addition, BERT already includes a CLS token in its original formulation, which is a special token prepended in front of every input sequence. Originally intended for sentence classification, the CLS token offers a convenient way for us to directly predict the target trajectory out of a predefined trajectory set.

\textbf{DistilBert} 
is a variant of BERT optimized for inference speed and model size. It is trained via knowledge distillation from BERT and retains 97\,\% of BERT's performance while reducing model size by 40\,\% \cite{sanh2019distilbert}.
Although, DistilBERT has been eclipsed in terms of performance by more recent models, its reduced size allows us to fine-tune the model for our trajectory prediction tasks on a single GPU without resorting to any specialized techniques. All variants of our architecture which process text-based input use DistilBERT.


\subsection{Joint Encoder}
\label{subsec:fused_model}

Given that the image and text encoders possess complementary strengths, we 
hypothesize that a model combining both encoders yields increased performance. 
In this work, we build a joint encoder by stacking the embeddings of the individual encoders.
We leave more complex joint encoders that process both input sources simultaneously as an exciting topic of future work.

%% file: sections/03c_representation.tex
\section{Scene Representation}
\label{sec:representation}

Next, we describe two different scene representations, rasterized images in Section \ref{subsec:image_representation} and text prompts in Section \ref{subsec:text_representation}. 
While the rasterized image presentation is routinely used for the AD prediction task, we are, to the best of our knowledge, the first to propose a text-based scene representation.
Our description contains the state of the agent for which predictions are made (target agent), its history and lane information.
Using a naive lane encoding exceeds the standard context length, and we propose a more efficient encoding based on Bézier curves instead.

\subsection{Image Representation}
\label{subsec:image_representation}

We  follow \cite{phan2020covernet} to generate a rasterized image representation of the scene. 
The scene is presented as an RGB image, where distinct colors are used to represent different semantic categories, such as the drivable area and crosswalks. 
To ensure consistent orientation, we rotate and translate the image such that the target agent is positioned at a fixed pixel location and its heading faces upwards.
We use a resolution of 0.1 meters per pixel and set the image height to 500 pixels and its width to 500 pixels. 
The agent's position is chosen to allow observation of the scene with a range of 40 meters ahead, 10 meters behind, and 25 meters to the left and right.

\subsection{Text Representation}
\label{subsec:text_representation}
Our approach is to construct a prompt for a language model which contains structured information about the traffic scene and the target agent.
Based on this prompt, the language model generates an encoding of the traffic scene which is used to predict the trajectory of the target agent from a set of fixed trajectory candidates.

Information about the target agent include the type of the vehicle, its current speed, acceleration, yaw rate and past $(x,y)$-positions. Since the coordinate system is centered around the current position of the target vehicle, the current position is always at $(0, 0)$ and this information is left out of the prompt.

In addition, we also include the following two components from the agent's environment:
a representation of the current lane the target agent is travelling on, and representations of surrounding lanes that the target agent could potentially switch to.
We further explore the following two options for lane representations:
(i)~a polyline representation where each lane is discretized at a resolution of 1\,meter and (ii)~a more compact representation based on Bézier curves that condense the representation into a fixed number of control points.

\textbf{Discretized Prompts}
contain lane information as a list of $(x,y)$-positions representing points on the lane sampled uniformly at 1\,meter intervals. This polyline representation is widely used by prediction algorithms working with vectorized map representations \cite{gao2020vectornet}.
However, the resulting prompts are on average 804 tokens long, and exceed the maximum context length of our language encoder of 512 tokens for 92\,\% of all samples. Prompts that are too long are truncated after 512 tokens, which leads to information loss, but allows the model to process the input.


\textbf{Bézier Prompts} offer a more compact alternative to discretized prompts.
Here, the lane is encoded  with Bézier curves, which rely on a set of discrete control points in order to generate parametric representations of smooth and continuous curves \cite{mortenson1999math4cg}.
In our experiments, we apply cubic Bézier curves, which are defined by four control points, the first and last of which coincide with the start and end point of the lane. The remaining two control points determine the curvature and direction and are fitted to minimize the mean-squared error between the Bézier curve and the discretized lane.

While not as widely used as the polyline representation, 
this representation allows for significantly shorter prompts (average length: 352) which, for all samples, fit into the context length of the language model without truncation. Figure~\ref{fig:prompt} shows an example for a Bézier prompt.





\begin{figure*}
    \centering
\fbox{\begin{minipage}{\textwidth}\footnotesize
You are an expert self-driving-car model, that can predict the future trajectory for a given vehicle, while also incorporating its current and past states, its current and possible future lanes and also information about other vehicles, pedestrians, drivable areas and other important sets of features. \\
            
\textbf{Task}: \\
Please predict the future trajectory for the given vehicle for the next 6 seconds, from a set number of fixed trajectories. \\

\textbf{Context Information}: \\
The 2D coordinate system (x,y) is from the prediction vehicle’s own frame of view.
Lane information is encoded as the 4 control points of a cubic Bezier curve. The first and last control point match with the beginning and end of the lane.

Prediction Vehicle: \\
Category: vehicle.car \\
Current Speed: 6.28[m/s] \\
Current Acceleration: 1.26[m/s²] \\
Current Yaw rate: 0.67[2$\pi$/s] \\
Past (x,y) positions in meters, sampled at 2 Hertz: \\
Time[s]    x[m]	y[m] \\
-2.0	      0.36	-11.63 \\
-1.5	      0.27	-8.8 \\
-1.0	      0.19	-5.97 \\
-0.5	      0.09	-3.15 \\

\textbf{Current Lane Information} (Bezier curve, as explained above): \\
x[m]	   y[m] \\
0.54	   -19.47 \\
0.53	   -6.59 \\
0.27	   6.32 \\
-0.23	   19.19 \\ 

\textbf{Possible Outgoing Lane Information} (Bezier curve, as explained above): \\
x[m]	    y[m] \\
-0.23	    19.19 \\
-0.71	    29.89 \\
-0.62	    40.59 \\
-0.98	    51.29 \\
x[m]	    y[m] \\
...  \\

Predicted trajectory number: 
\end{minipage}}
    \caption{\textbf{Example Prompt.} Our prompt contains information about the agent state, its history and lane information. 
    We use a compact lane encoding with the help of Bézier curves.}
    \label{fig:prompt}
\end{figure*} 

%% file: sections/04_experiments.tex
\section{Experiments}
\label{sec:experiments}

\begin{table*}
\begin{center}
\begin{tabular}{ l |c c c | c c c c c c c c}
Method & Image & Text & \#Modes & minADE$_1$ & minADE$_5$ & minADE$_{10}$ & MissRate$_1$ & MissRate$_5$ & MissRate$_{10}$ & minFDE$_1$\\ 
\hline
Constant velocity \& yaw & & & & 4.61 & 4.61 & 4.61 & 0.91 & 0.91 & 0.91 & 11.21\\  
Physics oracle & & &  & \textbf{3.70} & 3.70 & 3.70 & \textbf{0.88} & 0.88 & \textbf{0.88} & \textbf{9.09}\\
CoverNet (32M) &\checkmark & &64 & 5.16 & 2.41 & 2.18 & N/A  & 0.92 & N/A & 10.84\\
CoverNet (34M) &\checkmark & &415 & 5.07 & \textbf{2.31} & \textbf{1.76} &N/A & 0.83 & N/A & 10.62\\
CoverNet (41M) &\checkmark&& 2206 & 5.41 & 2.62 & 1.92 & N/A & \textbf{0.76} & N/A & 11.36\\
\hline
ResNet-50 (24M)&\checkmark& & 64 & 5.07 & 2.56 & 2.20 & 0.95 & 0.92 & 0.92 & 10.61\\
ResNet-50 (24M) &\checkmark&& 415 & 4.80 & 2.45 & 1.86 & 0.94 & 0.82 & 0.76 & 10.18\\
ResNet-50 (28M) &\checkmark&& 2206 & 5.31 & 2.80 & 2.11 & \textbf{0.93} & 0.76 & 0.64 & 11.22\\
ResNet-152 (58M) &\checkmark&& 64 & 4.86 & 2.47 & 2.17 & 0.95 & 0.92 & 0.92 & 10.15\\
ResNet-152 (59M) &\checkmark&& 415 & \textbf{4.51} & \textbf{2.33} & \textbf{1.80} & \textbf{0.93} & 0.81 & 0.76 & \textbf{9.57}\\
ResNet-152 (63M) &\checkmark&& 2206 & 4.72 & 2.58 & 1.94 & \textbf{0.93} & \textbf{0.75} & \textbf{0.63} & 10.05\\
\hline
BEiT-B (86M) &\checkmark&& 64 & 4.31 & 2.32 & 2.12 & 0.95 & 0.92 & 0.92 & 9.12\\
BEiT-B (86M) &\checkmark&& 415 & \textbf{3.92} & \textbf{1.98} & \textbf{1.57} & 0.92 & 0.79 & 0.74 & \textbf{8.46}\\
BEiT-B (88M) &\checkmark&& 2206 & 4.20 & 2.29 & 1.75 & \textbf{0.91} & \textbf{0.72} & \textbf{0.59} & 9.22\\
\hline
DistilBERT$_{discr.}$ (67M) &&\checkmark& 64 & 4.58 & 2.42 & 2.18 & 0.95 & 0.92 & 0.92 & 10.25\\
DistilBERT$_{discr.}$ (67M) &&\checkmark& 415 & 4.31 & 2.24 & 1.74 & 0.92 & 0.80 & 0.75 & 9.97\\
DistilBERT$_{discr.}$ (69M) &&\checkmark& 2206 & 4.86 & 2.80 & 2.11 & \textbf{0.91} & \textbf{0.70} & 0.57 & 11.30\\
DistilBERT (67M) &&\checkmark& 64 & 4.45 & 2.39 & 2.16 & 0.95 & 0.92 & 0.92 & 9.94\\
DistilBERT (67M) &&\checkmark& 415 & \textbf{4.23} & \textbf{2.20} & \textbf{1.70} & 0.93 & 0.80 & 0.75 & \textbf{9.81}  \\
DistilBERT (69M) &&\checkmark& 2206 & 4.56 & 2.55 & 1.94 & \textbf{0.91} & \textbf{0.70} & \textbf{0.56} & 10.57\\
\hline
BEiT-B-DistilBERT (159M) & \checkmark & \checkmark& 64 & 3.93 & 2.23 & 2.10 & 0.94 & 0.92 & 0.92 & 8.50\\
BEiT-B-DistilBERT (160M) & \checkmark & \checkmark & 415 & \textbf{3.62} & \textbf{1.87} & \textbf{1.49} & 0.92 & 0.78 & 0.73 & \textbf{8.09}\\
BEiT-B-DistilBERT (168M) & \checkmark & \checkmark & 2206 & 3.73 & 2.00 & 1.53 & \textbf{0.90} & \textbf{0.66} & \textbf{0.52} & 8.41
\end{tabular}
\end{center}
\caption{\textbf{Empirical Evaluation on nuScenes.} 
Our experiments show that our joint model outperforms individual text and image encoders, as well as baselines. 
Baselines are taken over from the CoverNet  paper \cite{phan2020covernet}. 
We highlight in bold the best performing method for each category and metric.
For all methods that build on the CoverNet decoder, we set the tolerance parameter to $\epsilon=8,4,2$ leading to $64, 412,2206$ modes (see Section \ref{sec:background}). 
We provide in brackets the number of parameters.
Models with subscript \emph{discr.} use discretized prompts with polyline lane representation.
}
\label{tab:results}
\end{table*}

Our experiments give a proof-of-concept that text encoders provide meaningful representations of scenes.
They complement image encoders, and their combination offers the best of both worlds leading to superior performance.

\subsection{nuScenes Dataset}
We evaluate our approach on the publicly available nuScenes dataset \cite{caesar2020nuscenes}, which is a common benchmark dataset in the autonomous driving community.
This dataset consists of 1,000 driving scenes recorded in Boston and Singapore.
Each scene is 20 seconds long and scenes are manually selected in order to ensure diversity.
The task is to predict the next 6-seconds long future trajectories based on its and neighboring agents’  2-seconds history and map information.

\subsection{Model Architectures}

\textbf{Unimodal architectures} use only a single input source. We contrast the performance of image-based encoders (ResNet, BEiT-B) against text-based encoders (DistilBERT) with discretized (disc.) and Bézier prompts. 
We directly use the Hugging Face Classifiers \cite{wolf2019huggingface} in our implementation.

\textbf{Joint architectures} combine text and image as input.  
We fuse the best-performing image encoder, BEiT-B, with our text encoder, DistilBERT.
Since the embeddings are frozen during fine-tuning of the joint model, we add an intermediate dense layer to blend the two modalities in the decoder.

\textbf{Baselines} consist of CoverNet \cite{phan2020covernet}  and the two physics-based models from its paper.

\subsection{Fine-Tuning}
\label{subsec:fine_tuning}

All models are initialized with pre-trained weights provided by Hugging Face~\cite{wolf2019huggingface} and fine-tuned on the nuScenes training dataset~\cite{caesar2020nuscenes}. 
Our unimodal architectures are fine-tuned by optimizing all weights of the model simultaneously. 

For the joint architecture, we freeze the fine-tuned weights of both, the image and text encoders, only optimizing the classification head.

We also conducted initial experiments with end-to-end fine-tuning of both encoders, but this approach was outperformed by the frozen model.
We think this behavior can be explained by each modality having its own convergence rate and shape, preventing them from converging simultaneously.
More complex training schemes will be explored in future work.

For all experiments, we report the results based on a single run. 
Within the run, we pick the best model based on the validation loss.



\subsection{Evaluation Metrics}
We apply the standard evaluation metrics that are provided in the \href{https://github.com/nutonomy/nuscenes-devkit}{nuScenes-devkit}: \textit{minimum Average Displacement Error} ($\text{minADE}_k$),
\textit{Final Displacement Error} ($\text{minFDE}_k$), and the miss rate over 2 meters, denoted as $\text{MissRate}_k$.
The subscript  $k=\{1,5,10\}$ denotes the number of the most probable predicted trajectories that are taken into account for metric calculation.
For all metrics, better performance is associated with smaller values.

\subsection{Empirical Analysis}
We present our results in Table \ref{tab:results}. 

\textbf{Images} are commonly used to represent the scene in the traffic forecasting domain. 
To ensure a fair comparison, we reimplemented a simplified version of the CoverNet architecture using directly the ResNet-50 classifier in Hugging Face.
Our simplified implementation achieves competitive results to the results reported in the CoverNet paper.
Moreover, we observe that switching from ResNet-50 (the backbone used in \cite{phan2020covernet}) to ResNet-152 further improves the predictive performance. 
These results are in line with the original ResNet publication \cite{he2016deep}, which also reports that model performance increases with depth.

Next, we switch the ResNet backbone with BEiT.
BEiT, a state-of-the-art vision Transformer architecture, demonstrates better predictive performance compared to ResNet.
Both, ResNet and BEiT are pre-trained on the ImageNet dataset \cite{russakovsky2015imagenet}. 
We believe that this performance gain can be attributed to the attention mechanism of BEiT, which enables it to extract more expressive features. 
This ultimately leads to improved generalization behavior.

\textbf{Text} was, in contrast to the image format, not used in prior work for scene representation.
Comparing the two different prompts, we observe that Bézier prompts outperform discretized prompts which can be most likely explained by
the fact that the discretized prompts are truncated for 92\% of all samples, while the Bézier prompts always fit into the context.

Comparing the text encoder with the image encoder, we find
that the best performing text-based architecture (DistillBERT with 2,206 modes) outperforms all image-based architectures in terms of miss rate. 
Contrarily,  the best performing image-based architecture (BEiT-B with 415 modes) outperforms all text-based architectures  in terms of average and final displacement error.
We find it remarkable that using a text only encoder achieves comparable results as the image-based encoder despite having less information available.
We hypothesize that text is more structured  than the image representation making it easier to extract information.
Furthermore, the inductive bias might be stronger for the text encoder than the image encoder: we directly provide outgoing lanes that can anchor the forecasts.

\textbf{Joint} representations, i.e., fusing text and image information, offer the best predictive performance compared to using a single modality only.
This indicates that both modalities have complementary strengths.

%% file: sections/05_summary.tex
\section{Summary}
\label{sec:summary}

In this paper, we  are the first to demonstrate the potential  of integrating pre-trained language models as text-based input encoders for the AD trajectory prediction task.
Our experiments confirm that text encoders are a valuable alternative to image encoders, 
and that joint encoders over both modalities perform better than using a single encoder in isolation.

While our experimental results are encouraging and our joint model significantly improves the baseline, it is important to acknowledge that its performance has not reached the state-of-the-art level yet (e.g \cite{deo2022multimodal}, \cite{gilles2022gohome}). 
Further optimization and model refinements are necessary.
Additionally, further analysis is required in order to understand the relative strengths and weaknesses of each encoder in different scenarios.
Nevertheless, we think that our study provides strong evidence that future research in this direction is needed.

%% file: sections/06_outlook.tex
\section{Outlook}
\label{sec:outlook}
Our contribution opens the door to a number of exciting opportunities for future investigations:

First, we think that our performance can be further improved by applying a different decoder. 
For simplicity, we have so far used the decoder of CoverNet \cite{phan2020covernet}, which is part of the \href{https://github.com/nutonomy/nuscenes-devkit}{nuScenes-devkit}.
While this decoder works by turning the trajectory prediction task into a classification task over a fixed set of trajectory candidates, recent papers suggest more expressive output representations such as a mixture of Gaussians \cite{look2023cheap} or a heatmap \cite{gilles2022gohome}.

Second, we have so far limited ourselves to models with less than 100\,M parameters. 
These models have the advantage that full fine-tuning is straight forward and can be performed on a single GPU without resorting to any specialized techniques.
However, the performance of language models increases with scale and current state-of-the-art models reach upward of 10\,B parameters (see e.g., \cite{touvron2023llama, dehghani2023scaling}).
Applying these large models in our setting could be an interesting way to boost performance, especially when combined with parameter-efficient fine-tuning \cite{hu2021lora} or soft prompting techniques \cite{lester2021softprompt}.
In addition, the increased context length of these models would allow us to provide additional input information such as the state of neighboring agents.

Third, when we fuse information from image and text representations, we first fine-tune each encoder in isolation, before concatenating their latent embeddings. 
While this approach already outperforms the versions of our model that use only one of the encoders, we expect an even larger performance gain when applying a joint image-and-text encoder such as \cite{li2023blip}. 
The joint encoder can capture and incorporate the inter-modal relationships between image and text more effectively, which may lead to improved feature representations.

Finally, language models in the AD domain are not restricted to the role of encoders in trajectory prediction tasks.
They can also be considered for generation of auxiliary textual output, e.g., by decoding the latent scene embedding into an explanation of the driving maneuver, which would enhance the prediction interpretability.
Another interesting direction is to apply language encoders for traffic simulation where scenario-specific instructions can guide the generation process \cite{tan2023language}.